\documentclass[10pt,journal,compsoc]{IEEEtran}
\usepackage{multicol}
\usepackage{multirow}
\usepackage{booktabs}
\usepackage{threeparttable}
\usepackage{array}
\usepackage{booktabs}
\usepackage{bbding}
\usepackage{diagbox}
 \usepackage{graphicx}
 \usepackage{amsmath}
 \usepackage{amsmath}
  \usepackage{amsfonts}
  \usepackage[mathscr]{euscript}
\ifCLASSOPTIONcompsoc
  \usepackage[nocompress]{cite}
\else
  \usepackage{cite}
\fi
\ifCLASSINFOpdf
\else
\fi
\begin{document}
%
\title{Intensity-Aware Loss for Dynamic Facial Expression Recognition in the Wild}
%
%
%
%

\author{Hanting Li, Hongjing Niu, Zhaoqing Zhu, 
and Feng~Zhao,~\IEEEmembership{Member,~IEEE}\\
 \thanks{H. Li, M. Sui, Z. Zhu, and F. Zhao are affiliated with University of Science and Technology of China, Hefei 230026, China (e-mail: fzhao956@ustc.edu.cn).}}
%
%

\markboth{Journal of \LaTeX\ Class Files,~Vol.~14, No.~8, August~2015}%
{Shell \MakeLowercase{\textit{et al.}}: Bare Demo of IEEEtran.cls for Computer Society Journals}
%



\IEEEtitleabstractindextext{%
\begin{abstract}
Compared with the image-based static facial expression recognition (SFER) task, the dynamic facial expression recognition (DFER) task based on video sequences is closer to the natural expression recognition scene. However, DFER is often more challenging. One of the main reasons is that video sequences often contain frames with different expression intensities, especially for the facial expressions in the real-world scenarios, while the images in SFER frequently present uniform and high expression intensities. However, if the expressions with different intensities are treated equally, the features learned by the networks will have large intra-class and small inter-class differences, which is harmful to DFER. To tackle this problem, we propose the global convolution-attention block (GCA) to rescale the channels of the feature maps. In addition, we introduce the intensity-aware loss (IAL) in the training process to help the network distinguish the samples with relatively low expression intensities. Experiments on two in-the-wild dynamic facial expression datasets (i.e., DFEW and FERV39k) indicate that our method outperforms the state-of-the-art DFER approaches. The source code will be made publicly available.
\end{abstract}
}
\maketitle
\IEEEdisplaynontitleabstractindextext

%
\IEEEpeerreviewmaketitle

\IEEEraisesectionheading{
\section{Introduction}}
Facial expressions are one of the most common human behaviors and play an important role in interpersonal communications \cite{darwin1998expression}. Facial expression recognition is the fundamental task of the intervention of mental diseases \cite{bisogni2022impact}, human-computer interaction \cite{liu2017facial}, and driver safety monitoring \cite{wilhelm2019towards}. In recent years, static facial expression recognition (SFER) has become a popular topic among affective computing tasks. Many SFER methods have gradually improved the accuracy of expression recognition based on a single image \cite{li2020deep,li2021mvt,she2021dive,wang2020SCN,xue2021transfer,zhao2021robust}. However, the facial expression is dynamic, and video-based data can provide richer facial information. Therefore, dynamic facial expression recognition (DFER) has gradually attracted more and more attentions. Unlike SFER, the DFER aims to classify a video into several discrete expressions (e.g., happiness, sadness, neutral, anger, surprise, disgust, and fear) \cite{jiang2020dfew}.

According to different data scenarios, the DFER datasets can be mainly divided into two categories: lab-controlled and in-the-wild. The lab-controlled datasets are collected in the laboratory settings, e.g., MMI \cite{pantic2005mmi} and CK+ \cite{lucey2010ck+}, so most sequences follow a relatively fixed pattern and have similar intensities (i.e., shift from a neutral facial expression to a peak expression). Researchers have proposed various techniques to effectively improve the performance of DFER methods in the laboratory scenarios \cite{yu2018dferlab,jeong2020ferlab}. Compared with lab-controlled DFER datasets, the in-the-wild ones are closer to the natural facial events and can provide more diverse data by collecting video sequences from the internet e.g., Aff-Wild \cite{zafeiriou2017affWILD}, DFEW \cite{jiang2020dfew} and FERV39k \cite{wang2022ferv39k}. As shown in Figure 1, the video sequences in the real world with different expression intensities could result in the problem that the inter-class distance becomes smaller than the intra-class distance. However, the supervised information of discrete-label datasets does not contain intensity-related priors such as DFEW and FERV39k, which leads to the fact that low-intensity expression sequences are often more likely to be misclassified. To address this issue, some datasets annotate video sequences with continuous labels along the arousal-valence axes, such as OMG \cite{barros2018omg} and AMIGOS \cite{miranda2018amigos}, and the arousal dimension can be used to measure the intensities of expressions. Nevertheless, annotating these datasets can be very expensive, and it is difficult to unify the annotation criteria of different annotators. Therefore, it is necessary to consider the impact of expression intensity on classifying low-cost datasets annotated by discrete labels.

\begin{figure}[t]
  \centering
  \includegraphics[width=8cm]{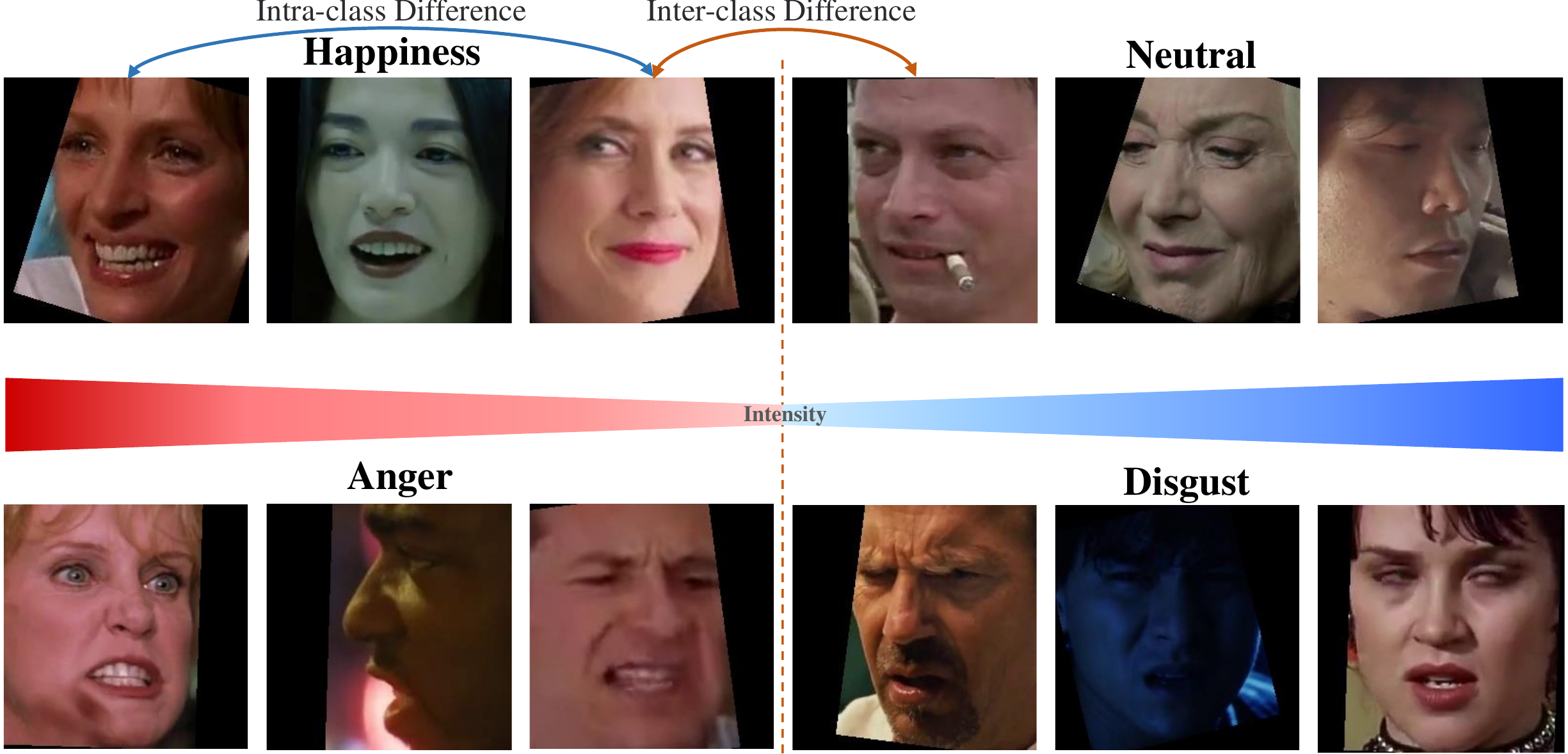}
  \caption{In-the-wild expression video sequences with different intensities. In both rows, the facial expression images in the middle have lower intensities, and the facial images on both sides have higher expression intensities.}
 
\end{figure}

For DFER in the wild, the early works are mainly designed based
on the hand-crafted features, such as LBP-TOP \cite{dhall2013LBPTOP}, STLMBP \cite{huang2014STLMBP}, and HOG-TOP \cite{chen2014HOG-TOP}. In recent years, with the development of parallel computing hardware and collection of large-scale DFER datasets \cite{wang2022ferv39k,jiang2020dfew}, deep-learning-based methods have gradually replaced the algorithms based on hand-crafted features and achieved state-of-the-art performance on the in-the-wild DFER datasets. 
Recently, vision transformer (ViT) \cite{dosovitskiy2020vit} has recently obtained promising results on many computer vision tasks, which inspires many researchers to build DFER models based on ViT \cite{zhao2021former-DFER,ma2022STT}. Since transformer has strong robustness against severe occlusion and disturbance \cite{naseer2021intriguing}, these transformer-based approaches mostly deal with various interferences in practical scenarios (e.g., variant head poses, poor illumination, and occlusions) by utilizing both spatial transformer and temporal transformer.

However, previous works on discrete-label DFER ignore the differences in the expression intensities between the samples of the same class (i.e., giving equal-status supervisory information to all video sequences). To solve this problem, we design a novel global convolution-attention block (GCA) to rescale the channels of the feature maps, which helps the network heighten critical channels in low-intensity samples (pull in the features of different-intensity sequences to reduce the intra-class differences) and suppresses less useful channels like other attention mechanisms do at the same time. Besides, we propose a novel intensity-aware loss (IAL) to help the network learn more information from hard-classified video sequences (i.e., low-intensity sequences).

In summary, this paper has the following contributions:
\begin{itemize}
	\item By aggregating information in each channel with a global convolution, we devise a plug-and-play global convolution-attention block that can rescale the channels of the feature maps. Specially, the GCA block can not only suppress the less important channel as done in other attention mechanisms (e.g., SE and CBAM), but also enhance the target-related channel for low-intensity expressions. To the best of our knowledge, this is the first work focusing on the problem of expression intensity in the discrete-label DFER task.
	\item We propose a simple but effective loss function called intensity-aware loss, which can force the network to pay extra attention to the most confusing class of low-intensity expression sequences. In turn, the network can learn more precise classification boundaries. 
	\item  Extensive ablation studies and visualization results demonstrate the effectiveness of our method. It outperforms the baseline model significantly and achieves state-of-the-art results on two popular in-the-wild DFER benchmarks.
\end{itemize}

\section{Related Work}
\subsection{DFER in the Wild}
Before the rise of deep learning, hand-crafted features occupied the mainstream in DFER, including LBP-TOP, STLMBP, and HOG-TOP. Besides, Liu et al. used different Riemannian kernels to measure the similarity/distance between sequences \cite{liu2014STM}. In recent years, deep-learning-based methods have gradually surpassed these methods using hand-crafted features, so we mainly discuss methods based on deep neural networks.

Unlike the SFER, which classify expressions based on a single image, DFER needs to explore the dependencies between frames of a video sequence. Therefore, an intuitive way is to add a network to model the temporal relation between the spatial features of frames. Under this paradigm, many methods \cite{lee2019rnn,lu2018rnn,ouyang2017rnn} use CNN-based models (e.g., ResNet \cite{he2016resnet} and VGG \cite{simonyan2014vgg}) to extract spatial features from each frame and then use RNN-based models such as LSTM \cite{hochreiter1997lstm} and GRU \cite{chung2014gru} to model the temporal dependencies between frames. Specially, Wang et al. introduced a multiple attention fusion network (MAFN) to fuse multimodal data by simulating human expression recognition mechanisms \cite{wang2019rnn}. Zhang et al. proposed an end-to-end STRNN to jointly integrate spatial and temporal dependencies \cite{zhang2018stRNN}. In addition to these ``two-stage'' methods, 3D-CNN-based methods adopt 3D convolution to extract spatial and temporal features jointly \cite{fan20163DCNN,lee20193DCNN,vielzeuf20173DCNN,hara20183dresnet}. Specially, Jiang devised a novel EC-STFL loss that boosted the performance of multiple 3DCNN and 2DCNN-based models \cite{jiang2020dfew}.

In recent years, with the introduction of transformer-based models from natural language processing tasks \cite{vaswani2017transformer} to computer vision tasks \cite{dosovitskiy2020vit}. Some researchers began using transformer to extract spatial and temporal features from the video sequences. Specifically, Zhao et al. devised a dynamic facial expression recognition transformer (Former-DFER) consisting of CS-Former and T-Former for learning spatial and temporal features, respectively \cite{zhao2021former-DFER}. Ma et al. proposed a spatio-temporal Transformer (STT), which can get the spatial and temporal information jointly by a transformer-based encoder \cite{ma2022STT}. Besides, Li et al. introduced a NR-DFERNet for suppressing the impact of noisy frames in video sequences \cite{li2022nrdfernet}.

Although the above methods have designed many effective models to tackle the discrete-label DFER problem, they all ignore the effect of the expression intensities on the DFER task. Therefore, we propose IAL to force the network to pay extra attention to expression sequences with low intensities.

\subsection{Attention Mechanisms}
\begin{figure*}[t]
  \centering
  \includegraphics[width=16.5cm]{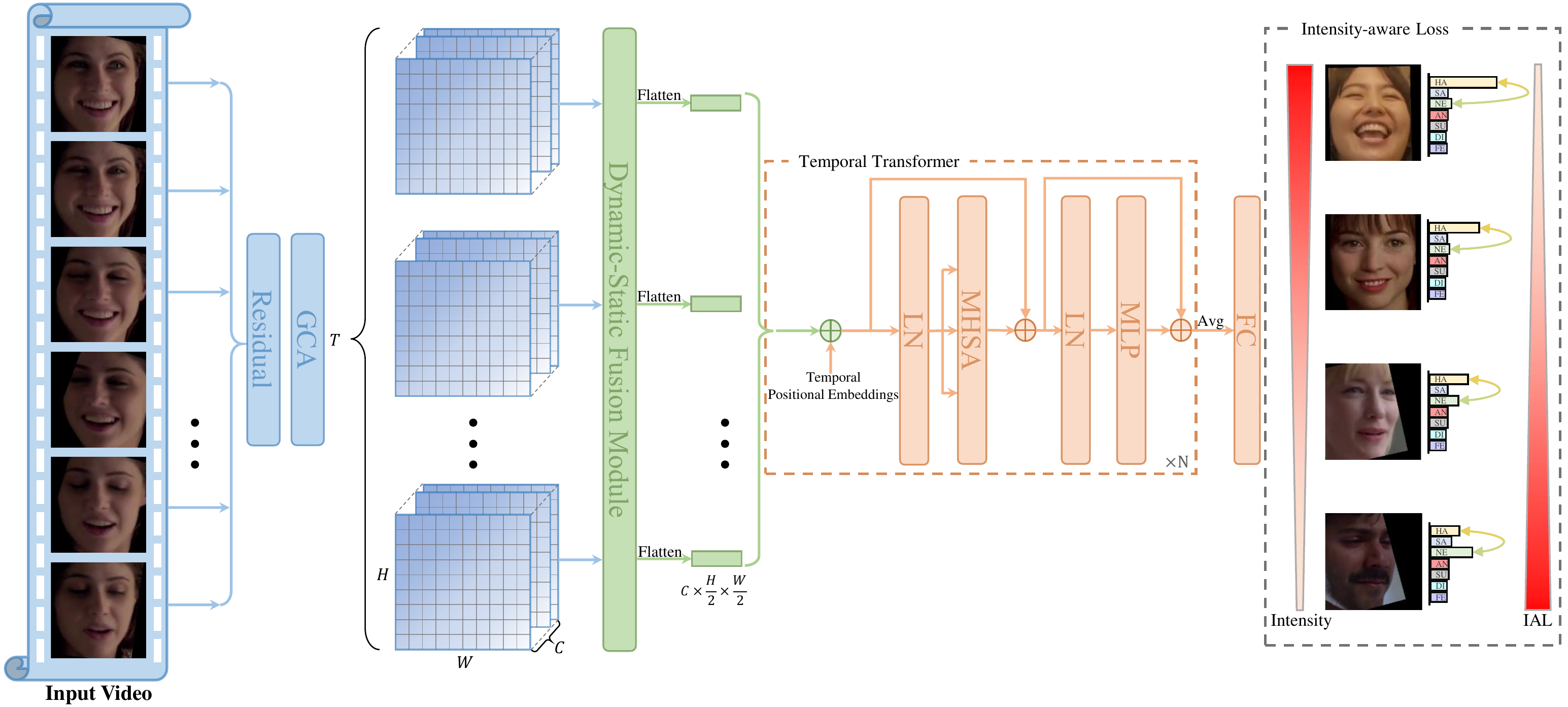}
  \caption{Overview of our method for DFER. The dynamic-static fusion module was proposed in \cite{li2022nrdfernet}. ``Residual'' stands for the building block of ResNet18 \cite{he2016resnet}. GCA and IAL represent the proposed global convolution-attention block and intensity-aware loss, respectively. LN, MHSA, and MLP separately denote the layer normalization, multi-heads self-attention, and multi-layer perceptron, while HA, SA, NE, AN, SU, DI, and FE individually indicates happiness, sadness, neutral, anger, surprise, disgust, and fear.}
\end{figure*} 
\begin{figure}[t]
  \centering
  \includegraphics[width=\linewidth]{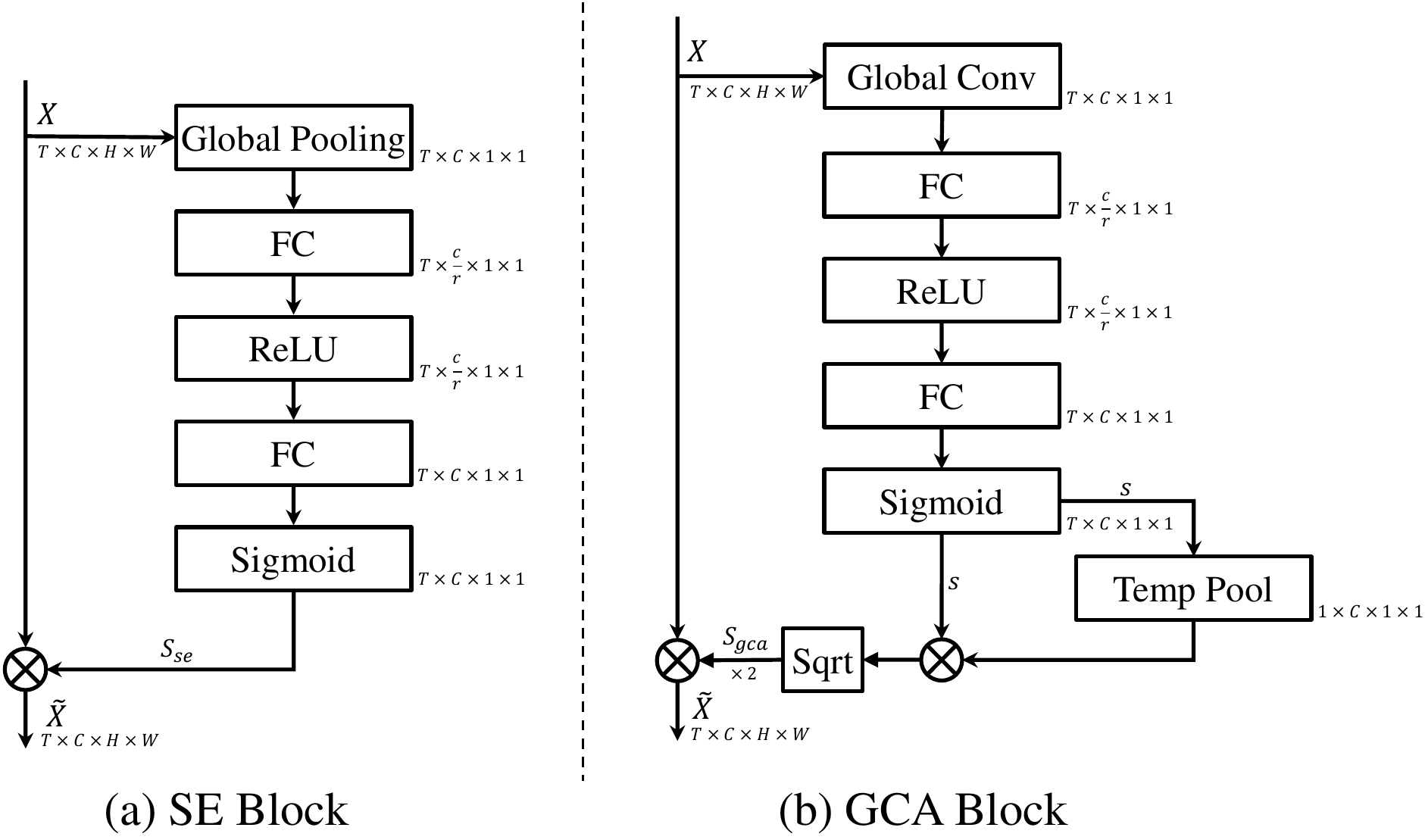}
  \caption{Diagram of the SE and proposed GCA blocks: (a) SE block and (b) our GCA block.}
\end{figure}
Attention mechanisms have been widely used in various visual tasks in recent years. By simulating the attention mechanism in the human vision system \cite{tootell1998attention}, the networks can focus on task-relevant information and suppress task-irrelevant information. Among them, SE block \cite{hu2018SE} squeezed global spatial information into a channel descriptor by doing global average pooling operation and modeled channel-wise relationships using two fully connected (FC) layers. ECA-Net \cite{wang2020ECA} replaced the FC layer in the SE block with a 1-D convolution filter to reduce the model complexity. In addition to modeling channel-wise dependencies, CBAM \cite{woo2018CBAM} learned a spatial-attention mechanism. Similarly, SGE \cite{li2019SGE} also combined spatial and channel attention. Specially, DANet \cite{fu2019DANet} utilized a two-branch structure to learn spatial attention and channel attention separately. 

Most of the above attention mechanisms use global spatial pooling (e.g., global average pooling or global max pooling) to aggregate channel information before learning the channel dependencies and then use the sigmoid function to constrain the channel-attention weights. 

However, the global pooling can lose spatial location information, which is vital for FER (e.g., the wrinkle-related features extracted from eyebrows or mouth can make a big difference to results). Besides, since the value range of the sigmoid function belongs to $(0, 1)$, the generated channel-attention weights can only be used to suppress useless channels, which does not meet our original intention of enhancing the channel features of low-intensity samples. Based on the above two reasons, we propose a global convolution-attention block to model the channel dependencies and enhance the critical channels.

\section{Method}
\subsection{Overview}
As shown in Figure 2, the facial expression sequence with the length of $T$ is dynamically sampled from the raw video as an input. Then the input clips $X_{in}$$\in$$\mathbb{R}^{T\times 3\times H_{in}\times W_{in}}$ are fed into several building blocks \cite{he2016resnet} to extract the frame-level features. Then the proposed GCA block is used to rescale the channels of the
feature maps (i.e., suppress the less important channel and enhance the target-related channel). Then the dynamic-static fusion module proposed in \cite{li2022nrdfernet} is applied for learning both static features of each frame and dynamic features between adjacent frames. The frame-level features that incorporate both dynamic and static features are flattened before being fed into a temporal transformer to learn the long-distance dependencies between frames. Subsequently, the mean value of the token sequence is fed into a FC layer to obtain the recognition result. Both the proposed IAL and cross-entropy loss are used to optimize the network.

\subsection{Input Clips}
Our method takes a clip $X_{in}$ consisting of $T$ RGB images as input. The input clips are dynamically sampled from the raw video as did in previous works. Specifically, for the training clips, the raw sequences are divided into $U$ segments equally and then randomly pick $V$ frames from each segment. As for the test clips, we first split all frames into U segments and then select V frames in the mid of each segment. Therefore, the length of the sampled clip is $T = U \times V$ for both the training set and testing set.

\subsection{Global Convolution-attention Block}
SE block \cite{hu2018SE} was proposed to learn the channel-wise dependencies of the features. As shown in Figure 3(a), the attention weights $S_{se}$$\in$$(0,1)$ generated by SE block are used to suppress the less important channels. For an input $X$$\in$$\mathbb{R}^{T\times C \times H \times W}$, where $T$ is the number of frames, $C$ is the number of the channels, while $H$ and $W$ are the height and width of the feature, respectively. The output $\tilde{X}$ of SE block can be formulated as,
\begin{equation}\label{1}\small
\tilde{X} =S_{se} \otimes X,
\end{equation}
\noindent with
\begin{equation}\label{3}\small
S_{se}= \sigma (W_{2} \delta (W_{1} Z)),
\end{equation}
\begin{equation}\label{2}\small
Z_{c}=\frac{1}{H\times W} \sum_{i=1}^{H} \sum_{j=1}^{W} X_{c}(i,j).
\end{equation}
\noindent Where $Z$$\in$$\mathbb{R}^{C}$ stands for the channel descriptor generated by a global average pooling operation on input $X$, and $Z_{c}$ and $X_{c}$ are the $c$-th element of $Z$ and $X$, respectively. $\delta$ refers to the ReLU function, and $\sigma$ represents the sigmoid function. $W_{1}$$\in$$\mathbb{R}^{\frac{C}{r}\times C}$ and $W_{2}$$\in$$\mathbb{R}^{C\times \frac{C}{r}}$ are the weights of the first and the second FC layers, respectively. $r$ denotes a dimensionality-reduction ratio, and $\otimes$ is a broadcast element-wise multiplication.

However, since the value of $S_{se}$ belongs to $(0,1)$, the SE block can only suppress the less important channel. A suitable attention mechanism for DFER is also supposed to enhance the key channels, which is crucial for strengthening the target-related features of low-intensity samples. Besides, location information is also critical for the DFER task (e.g., wrinkle-related features appearing on the corners of the mouth or between the eyebrows can indicate a completely different expression), but the channel descriptor $Z$ generated by global average pooling losses the location information of the input features.

Based on the above viewpoints, we propose the GCA block as shown in Figure 2(b). For a same input $X$$\in$$\mathbb{R}^{T\times C \times H \times W}$, the output of GCA block can be calculated by, 

\begin{equation}\label{4}\small
\tilde{X} =S_{gca} \otimes X,
\end{equation}
\noindent with
\begin{equation}\label{6}\small
S_{gca}= 2\sqrt{s\otimes GAP_{t}(s)},
\end{equation}
\begin{equation}\label{7}\small
s= \sigma (W_{2} \delta (W_{1} Z)),
\end{equation}
\noindent and 
\begin{equation}\label{5}\small
Z_{c}=\sum_{i=1}^{H} \sum_{j=1}^{W} X_{c}(i,j)\times \hat{W}_{c}(i,j).
\end{equation}
\noindent Where $S_{gca}$ is the attention weights generated by GCA block, and $\hat{W}_{c}$$\in$$\mathbb{R}^{H \times W}$ stands for the global convolution kernel of the $c$-th channel. $GAP_{t}$ denotes a global average pooling on temporal dimensions.

Since $S_{gca}$ belongs to $(0,2)$, the GCA block can not only inhibit less important channels but also enhances critical channels. At the same time, the global convolution can retain the location information of the features, which is essential for DFER. Besides, a temporal global average pooling offers a larger inter-frame field of view.

\subsection{Intensity-aware Loss}
Expression intensity is an important attribute of human facial expressions. When we define $\rm{intensity}$$\in$$(0,1)$, it is obvious that all non-neutral expressions tend to approach neutral expressions when the intensity converges to zero, which can be defined as,
\begin{equation}\label{8}\small
\lim_{\rm{intensity} \to 0} NNE=NE.
\end{equation}
\noindent Where NNE and NE are the non-neutral and neutral expressions, respectively. It seems that taking facial expressions as a regression task is more appropriate. However, since DFER datasets with continuous labels are expensive to annotate, the scale of such datasets is often limited. Besides, the intensity criteria for annotating are difficult to unify. To tackle this, we propose intensity-aware loss to reduce the effect brought by low-intensity expression samples to the discrete-label DFER.

Based on the assumption that a low-intensity sample is likely to be confused with low-intensity samples from other classes (as shown in Figure 1), the network is supposed to pay extra attention to the most confusing category of each sample. Therefore, the proposed intensity-aware loss can be formulated as,
\begin{equation}\label{9}\small
 \mathscr{L}_{IA}=-\rm{log}(\it{P_{IA}}),
\end{equation}
\noindent with
\begin{equation}\label{10}\small
P_{IA}=\frac{e^{x_t}}{e^{x_t}+e^{x_{max}}}.
\end{equation}
\noindent Where $x_t$ denotes the logits of the target class and $x_{max}$ stands for the largest logits excluding the target class. In summary, the proposed IAL gives extra attention to the class that is most likely to cause confusion as shown in Figure 2. Finally, with the commonly used cross-entropy loss, the overall loss function is defined as follows,
\begin{equation}\label{11}\small
 \mathscr{L}=\mathscr{L}_{CE}+\lambda\mathscr{L}_{IA},
\end{equation}
\noindent where $\lambda$ is a hyper-parameter controlling
loss coefficients. It is worth noting that when $\mathscr{L}_{CE}$ converges to zero, $\mathscr{L}_{IA}$ also converges to zero.
\section{Experiments}
We carry out extensive experiments on two popular in-the-wild DFER datasets (i.e., DFEW \cite{jiang2020dfew} and FERV39k \cite{wang2022ferv39k}).It is worth noting that FERV39k has just been released in 2022 and is currently the largest publicly available in-the-wild DFER dataset. In this section, we first introduce the datasets and implementation details. Then we explore the effectiveness of GCA block and intensity-aware loss. Subsequently, we compare the proposed method with several SOTAs and give some visualizations to demonstrate the effectiveness of our method.
\subsection{Datasets}
\textbf{DFEW} \cite{jiang2020dfew} consists of 16372 video clips from more than 1500 movies. These video clips contain various challenging interferences in practical scenarios such as extreme illumination, occlusions, and capricious pose changes. In addition, each video clip is individually annotated by ten well-trained annotators under professional guidance and assigned to one of seven basic expressions (i.e., happiness, sadness, neutral, anger, surprise, disgust, and fear). Consistent with the previous works, we only conduct experiments on 12059 video clips, which can be clearly assigned to a specific single-labeled emotion category. All the samples have been split into five same-size parts (fd1\textasciitilde fd5) without overlap. We choose 5-fold cross-validation, which takes one part of the samples for testing and the remaining for training, as the evaluation protocol. For all the experiments, the unweighted average recall (UAR) and weighted average recall (WAR) are used for evaluations.

\textbf{FERV39k} \cite{wang2022ferv39k} is current the largest in-the-wild DFER dataset and contains 38935 video clips collected from 4 scenarios, which can be subdivided into 22 fine-grained scenes (e.g., daily life, talk show, business, and crime). Each video clip is independently annotated by 30 professional annotators to ensure high-quality labels and assigned to one of the seven primary expressions as with DFEW. In FERV39k Benchmark, video
clips of all scenes are randomly shuffled and split into training (80\%/31088 clips) and testing (20\%/7847 clips) sets without overlapping. Therefore for a fair comparison, we directly use the training set and the testing set divided by FERV39k. UAR and WAR are also deployed as the evaluation metrics.

\subsection{Implementation Details}
\textbf{Training Setting:} In our experiments, all facial images are resized to the size of 112 $\times$ 112. Random cropping, horizontal flipping, and color jittering are employed to avoid over-fitting. We use SGD \cite{robbins1951sgd} to optimize our model with a batch size of 40. For the DFEW dataset, the learning rate is initialized to 0.001 and decreased at an exponential rate in 80 epochs for intensity-aware and cross-entropy loss function. For the FERV39k dataset, the learning rate is also initialized to 0.001, decreased at an exponential rate in 100 epochs for intensity-aware and cross-entropy loss function. For both datasets, models are trained from scratch. As for sampling, the length of the dynamically sampled sequence is 16 ($U$= 8, $V$= 2 for FERV39K and $U$= 16, $V$= 1 for DFEW). The number of the self-attention heads and the temporal transformer encoders are set at 4 and 2. By default, the dimensionality-reduction ratio $r$ is set at 16, and the loss coefficients $\lambda$ is set at 0.1. All the experiments are conducted on a single NVIDIA RTX 3090 card with PyTorch toolbox \cite{paszke2019pytorch}.

\textbf{Validation Metrics:} Consistent with previous methods, we take the unweighted average recall (UAR, i.e., the accuracy per class divided by the number of classes without considering the number of instances per class) and weighted average recall (WAR, i.e., accuracy) as the metrics.
\begin{table}\small
\begin{center}

\label{table:headings}\caption{Evaluation of the proposed GCA block and intensity-aware loss. The best results are in bold.}
\begin{tabular}{c|cc|cc|cc}
\toprule
\multirow{2}{*}{Setting}&\multicolumn{2}{c|}{Methods}&\multicolumn{2}{c|}{DFEW (\%)}&\multicolumn{2}{c}{FERV39k (\%)}\cr
    \cmidrule(lr){2-7}&GCA&IAL& UAR & WAR & UAR & WAR\cr
\midrule
a &              &               &53.56 &67.42 &32.79&45.24\cr
b& \CheckmarkBold&                  &54.12 &68.03&34.67&46.90  \cr
c&              & \CheckmarkBold &54.58 &68.57 &34.96&47.66\cr
d&\CheckmarkBold &\CheckmarkBold   &\textbf{55.71} &\textbf{69.24}&\textbf{35.82}&\textbf{48.54}  \cr
\bottomrule
\end{tabular}
\end{center}
\end{table}
\begin{table}\small
\begin{center}

\caption{Comparison of the proposed GCA block, SE block \cite{hu2018SE}, and channel-attention module of CBAM block \cite{woo2018CBAM}. The auxiliary classifier is used by default. ``w/o aux'' means no auxiliary classifier is used to optimize the network. The best results are in bold.
}
\begin{tabular}{c|cc|cc}
\toprule
\multirow{2}{*}{Methods}&\multicolumn{2}{c|}{DFEW (\%)}&\multicolumn{2}{c}{FERV39k (\%)}\cr
    \cmidrule(lr){2-5}
    & UAR & WAR& UAR & WAR\cr
\midrule
SE (w/o aux)  &53.10 &66.98&32.85&45.56 \cr
CBAM (w/o aux)&53.31&67.14 &33.41&45.76 \cr
GCA (w/o aux)&53.97 &67.74&34.11&46.45\cr
\midrule
SE   &53.47 &67.32&33.16&45.90 \cr
CBAM &53.78&67.64&33.89&46.15 \cr
GCA &\textbf{54.12}&\textbf{68.03}&\textbf{34.67} &\textbf{46.90}\cr
\bottomrule
\end{tabular}\label{table:headings}
\end{center}
\end{table}

\begin{table}[t]\small
\begin{center}

\caption{Evaluation of different dimensionality-reduction ratios $r$ and loss coefficients $\lambda$. The best results are in bold.
}

\begin{tabular}{c|c|cc|cc}
\toprule
\multicolumn{2}{c|}{Setting}&\multicolumn{2}{c|}{DFEW (\%)}&\multicolumn{2}{c}{FERV39k (\%)}\cr
    \cmidrule(lr){1-6}$\lambda$&$r$
    & UAR & WAR& UAR & WAR\cr
\midrule
0.1 &8 &54.58& 68.64                           &35.08& 47.60\cr
0.15 &8 &54.43& 68.57                           &35.27&47.79  \cr
0.3 &8 &54.16&  68.19                         &35.02&47.47\cr
0.1 &16 &\textbf{55.71}&\textbf{69.24}&35.56&48.42\cr
0.15 &16 &55.24& 69.02                    &\textbf{35.82}&\textbf{48.54} \cr
0.3 &16 &54.82& 68.76               &35.21 &48.11\cr
\bottomrule
\end{tabular}\label{table:headings}
\end{center}
\end{table}
 
\subsection{Ablation Studies}
We conduct ablation studies on DFEW and FERV39k to demonstrate the effectiveness of each component in our method (i.e., GCA and IAL). 
We take the model consisting of building blocks of ResNet18 \cite{he2016resnet}, dynamic-static fusion module of NR-DFERNet \cite{li2022nrdfernet} and vanilla transformer \cite{dosovitskiy2020vit,vaswani2017transformer} as a baseline in our experiments (i.e., setting (a) in Table 1). By default we add auxiliary classifiers connected to the intermediate attention blocks (e.g., SE, CBAM, and GCA) as did in \cite{szegedy2015googlenet} to help the attention mechanism generate attention weights according to the target then pull in the features of different-intensity samples.

\textbf{Evaluation of GCA Block and Intensity-Aware Loss:} We first study the effectiveness of GCA and IAL. As shown in Table 1, replacing the building block used in ResNet18 with the proposed GCA block (setting (b)) and introducing IAL to force the network to pay extra attention to the easily-confused class (setting (c)) can both boost the overall performance on two benchmarks. Specifically, when we deploy both GCA and IAL, our method setting (d) exceeds the baseline setting (a) by 2.15\%/1.82\% of UAR/WAR on DFEW and 3.03\%/3.30\% of UAR/WAR on FERV39k, which fully indicates effectiveness of the proposed modules.

\textbf{Comparison with Other Channel-Attention Modules:} We compare our GCA block with several classic channel-attention modules in Table 2. In addition, we also explore the effectiveness of the auxiliary classifiers \cite{szegedy2015googlenet}. It can be seen that the performance of all attention mechanisms is improved after deploying auxiliary classifiers to generate attention weights based on the target label. Besides, among different channel-attention blocks, our GCA block exceeds both SE block and CBAM block on two benchmarks. Specifically, GCA outperforms SE by 0.65\%/0.71\% of UAR/WAR on DFEW and 1.51\%/1.00\% of UAR/WAR on FERV39k. The main advantage of GCA is that the location information of features is preserved by replacing global pooling with global convolution.

\textbf{Evaluation of Hyper-Parameters:} We conduct ablation studies on several key hyper-parameters (i.e., the loss coefficients $\lambda$ of IAL and the dimensionality-reduction ratio $r$ of GCA block) of our method to explore their impact on performance. As shown in Table 3, smaller loss coefficients tend to have a better performance, whose reason is that a large loss coefficient will make the network pay too much attention to the most confusing category and ignore the distinction between other categories. As for the dimensionality-reduction ratio, larger $r$, which corresponds to smaller parameter sizes, tends to have a better performance on two datasets. In summary, all the results in Table 3 are significantly improved compared to the baseline in Table 1, which also shows the robustness of our method.

\subsection{Comparison with State-of-the-Arts}
In this section, we compare our best results with current state-of-the-art methods on the DFEW and FERV39k benchmarks to demonstrate the superiority of our methods.

\textbf{Results on DFEW:} Consistent with the previous works, the experiments are conducted under 5-fold cross-validation. The experimental results are shown in Table 4. As can be seen, our methods achieve the best results both in WAR and UAR. Specifically, the proposed method outperforms NR-DFERNet \cite{li2022nrdfernet} by 1.5\%/1.05\% of UAR/WAR. We also show the performance on each expression in Table 4. It is obvious that ``fear'' and ``disgust'' get poorer results than other classes, which are mainly on account of the insufficient training samples (the proportion of “disgust” and “fear”
is 1.22\% and 8.14\%, respectively).

\textbf{Results on FERV39k:} As the current largest in-the-wild DFER dataset, FERV39k benchmark contains diverse data sources (22 fine-grained scenes), which poses challenges for DFER tasks. As shown in Table 5, our method significantly outperforms the compared methods in WAR and achieves comparable results with Former-DFER \cite{zhao2021former-DFER} in UAR. Specifically, the proposed method exceeds Former-DFER by 1.69\% of WAR. It should be noticed that FERV39k also has a imbalanced data distribution. The proportions of “disgust” and “fear” sequences are 5.89\% and 5.4\%, which is the reason why our method achieve a relatively low performance in UAR.
\begin{figure}[t]
  \centering
  \includegraphics[width=\linewidth]{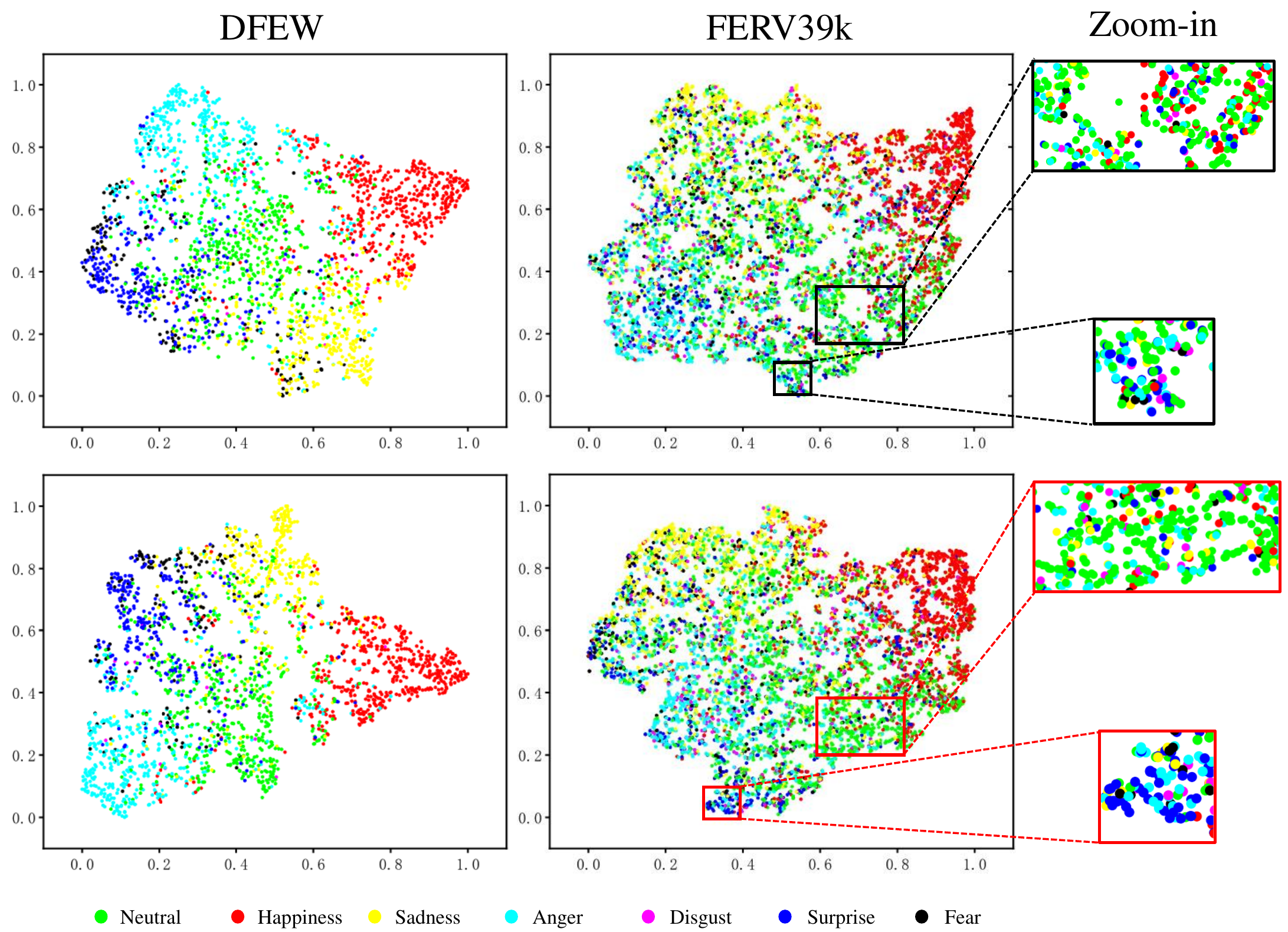}
  \caption{Visualization of the feature distribution (t-SNE \cite{van2008tsne}) learned by the baseline (top) and our method (bottom) on two datasets.}
\end{figure}
\subsection{Visualization}
 We utilize t-SNE \cite{van2008tsne} to analyze the feature distribution learned by the baseline and our methods in Figure 4. 
 \begin{table*}[t]\small
\begin{center}
\caption{Comparison with state-of-the-art methods on DFEW. $^\dagger$ denotes time interpolation \cite{zhou2011ti,zhou2013ti}. $^\ddagger$ indicates dynamic sampling used in \cite{zhao2021former-DFER,ma2022STT,li2022nrdfernet}. The bold denotes the best.
}
\begin{tabular}{c|ccccccc|cc}
\toprule
\multirow{2}{*}{Method}&
    \multicolumn{7}{c|}{Accuracy of Each Emotion (\%) }&\multicolumn{2}{c}{ Metrics (\%)}\cr
    \cmidrule(lr){2-10}&Happiness & Sadness & Neutral & Anger & Surprise & Disgust & Fear & UAR & WAR\cr
\midrule
C3D$^\dagger$ \cite{tran2015C3D}  &75.17 &39.49 &55.11 &62.49 &45.00 &1.38 &20.51 &42.74 &53.54\cr
P3D$^\dagger$ \cite{qiu2017p3d}  &74.85 &43.40 &54.18 &60.42 &50.99 &0.69 &23.28 &43.97 &54.47\cr
R(2+1)D18$^\dagger$ \cite{tran2018closer}  &79.67 &39.07 &57.66 &50.39 &48.26 &3.45 &21.06 &42.79 &53.22\cr
3D Resnet18$^\dagger$  &73.13 &48.26 &50.51 &64.75 &50.10 &0.00 &26.39 &44.73 &54.98 \cr
I3D-RGB$^\dagger$ \cite{carreira2017I3D-RGB} &78.61 &44.19 &56.69 &55.87 &45.88 &2.07 &20.51 &43.40 &54.27\cr
VGG11+LSTM$^\dagger$ &76.89 &37.65 &58.04 &60.70 &43.70 &0.00 &19.73 &42.39 &53.70 \cr
ResNet18+LSTM$^\dagger$  &78.00 &40.65 &53.77 &56.83 &45.00 &\textbf{4.14} &21.62 &42.86 &53.08 \cr
3D R.18+Center Loss$^\dagger$ \cite{wen2016discriminative} &78.49 &44.30 &54.89 &58.40 &52.35 &0.69 &25.28 &44.91 &55.48 \cr
EC-STFL$^\dagger$ \cite{jiang2020dfew} &79.18 &49.05 &57.85 &60.98 &46.15 &2.76 &21.51 &45.35 &56.51 \cr
3D Resnet18$^\ddagger$  &76.32 &50.21 &64.18 &62.85 &47.52 &0.00 &24.56 &46.52 &58.27 \cr
ResNet18+LSTM$^\ddagger$  &83.56 &61.56 &68.27 &65.29 &51.26 &0.00 &29.34 &51.32 &63.85 \cr
Resnet18+GRU$^\ddagger$  &82.87 &63.83 &65.06 &68.51 &52.00 &0.86 &30.14 &51.68 &64.02 \cr
Former-DFER$^\ddagger$ \cite{zhao2021former-DFER} &84.05 &62.57 &67.52 &70.03 &56.43 &3.45 &31.78 &53.69 &65.70 \cr
STT$^\ddagger$\cite{ma2022STT}&87.36&67.9&64.97&71.24&53.1&3.49&\textbf{34.04}&54.58&66.45\cr
NR-DFERNet$^\ddagger$ \cite{li2022nrdfernet}&\textbf{88.47}&64.84&70.03&75.09&61.60&0.00&19.43&54.21&68.19\cr

\midrule
GCA+IAL$^\ddagger$ (Ours)&87.95&\textbf{67.21}&\textbf{70.10}&\textbf{76.06}&\textbf{62.22}&0.00&26.44&\textbf{55.71}&\textbf{69.24}\cr
\bottomrule
\end{tabular}\label{table:headings} 
\end{center}
\end{table*}

\begin{table}[t]\small
\begin{center}

\caption{Comparison with state-of-the-art methods on FERV39k. All experiments use the dynamic sampling strategy. The bold denotes the best.
}
\begin{tabular}{c|cc}
\toprule
\multirow{2}{*}{Method}&\multicolumn{2}{c}{ Metrics (\%)}\cr
    \cmidrule(lr){2-3}& UAR & WAR\cr
    
\midrule
C3D \cite{tran2015C3D}  &22.68 &31.69 \cr
P3D \cite{qiu2017p3d} & 23.20 &33.39\cr
I3D-RGB \cite{carreira2017I3D-RGB} &30.17 &38.78 \cr
R(2+1)D \cite{tran2018closer} & 31.55& 41.28\cr
3D ResNet18  &26.67 &37.57 \cr
VGG13+LSTM  &32.41 &43.37 \cr
VGG16+LSTM  &30.43 &41.47 \cr
ResNet18+LSTM  &30.92 &42.95 \cr
Two VGG13+LSTM  &32.79 &44.54 \cr
Two ResNet18+LSTM  &31.28 &43.2 \cr
Former-DFER \cite{zhao2021former-DFER} &\textbf{37.20} &46.85 \cr
NR-DFERNet \cite{li2022nrdfernet}&33.99&45.97\cr
\midrule
GCA+IAL (Ours) &35.82&\textbf{48.54}\cr
\bottomrule
\end{tabular}\label{table:headings}

\end{center}
\end{table}
 For the DFEW dataset, the features of our method are more closely distributed, and the boundaries between the different classes are also more pronounced. It should be noticed that the features of neutral sequences learned by the baseline are distributed at the center of the remaining categories, which is consistent with our analysis in Eq. (8) (i.e., non-neutral expressions tend to approach neutral expressions when the intensity converges to zero). In comparison, the features learned by our method have more apparent boundaries between neutral and non-neutral samples. Especially for neutral sequences, the features of neutral samples are no longer at the center of non-neutral expression features, which can be explained as the features of low-intensity sample are enhanced by our method and no longer converge uniformly to neutral expressions. As for the FERV39k dataset, it can be seen that our method learns more discriminative features than the baseline, especially for the ``neutral'' and ``surprise'' samples. Since the overall performance on FERV39k is still weak, the t-SNE result still looks a little bit messy. Therefore, we zoom in on critical areas to give a more intuitive display.
 
\subsection{Discussions}
Whether from the visualization results in Figure 4 or the images in Figure 1, we can find the challenges that low-intensity samples pose to the discrete-label DFER, which is entirely different from the static expression recognition task based on a single image with high intensity. Therefore, we first raise this concern on discrete-label DFER and propose IAL to find these low-intensity samples, and then use the proposed GCA block to enhance their features.

However, there is still much room for improvement in accurately discovering these low-intensity samples and enhancing their features without additional supervision information. We hope this work can inspire researchers to propose more effective solutions to this issue.

\section{Conclusion}

In this paper, we develop a plug-and-play module called global convolution-attention block and a simple but effective intensity-aware loss for in-the-wild DFER. Our global convolution-attention block is designed to rescale the channels of feature maps, so that the target-related features of low-intensity sequences can be enhanced and the less important features are suppressed. The proposed intensity-aware loss helps the networks pay extra attention to the most likely confusing class of each sample (low-intensity samples often have a likely confusing class). The experiments and visualization results demonstrate the effectiveness and superiority of our method.

Specifically, we focus on dealing with the low-intensity samples in in-the-wild dynamic facial expression recognition task. To the best of our knowledge, this is the first work focusing on the expression intensity problem in discrete-label DFER. We hope that more researchers can note this issue and provide more interesting solutions in the future.


%



  

\ifCLASSOPTIONcaptionsoff
  \newpage
\fi



%

\bibliographystyle{IEEEtran}
\bibliography{bare_jrnl_compsoc}

\end{document}